\documentclass[conference]{IEEEtran}
\IEEEoverridecommandlockouts

\usepackage{amsmath,amssymb,amsfonts}
\usepackage{graphicx}
\usepackage{booktabs}
\usepackage{array}
\usepackage{balance}
\usepackage{newtxtext,newtxmath}
\usepackage{tikz}
\usetikzlibrary{arrows.meta,positioning,calc,decorations.pathreplacing,patterns}
\usepackage{pgfplots}
\pgfplotsset{compat=1.18}

\newtheorem{remark}{Remark}
\newtheorem{assumption}{Assumption}
\newtheorem{proposition}{Proposition}
\newtheorem{definition}{Definition}
\newtheorem{lemma}{Lemma}
\newtheorem{corollary}{Corollary}
\newtheorem{theorem}{Theorem}

\newcommand{\bv}[1]{\boldsymbol{#1}}
\newcommand{\mat}[1]{\mathbf{#1}}
\newcommand{\cF}{\mathcal{F}}
\newcommand{\cA}{\mathcal{A}}

\makeatletter
\renewenvironment{abstract}{\normalfont\small\bfseries\noindent
  \textit{Abstract}\ }{\par\vspace{0.7\baselineskip}}
\renewenvironment{IEEEkeywords}{\normalfont\small\bfseries
  \vspace{0.4\baselineskip}\noindent\textit{Index Terms}\ }
  {\par\vspace{0.6\baselineskip}}
\makeatother

\newcommand{\cS}{\mathcal{S}}
\newcommand{\cK}{\mathcal{K}}
\newcommand{\cW}{\mathcal{W}}
\newcommand{\cC}{\mathcal{C}}
\newcommand{\T}{\mathrm{T}}
\newcommand{\sgn}{\operatorname{sgn}}
\newcommand{\dist}{\operatorname{dist}}

\makeatletter
\renewcommand{\section}{\@startsection{section}{1}{\z@}%
  {1.0ex plus 1ex minus 0.3ex}{0.45ex plus 0.4ex minus 0ex}%
  {\normalfont\normalsize\centering\scshape}}
\renewcommand{\subsection}{\@startsection{subsection}{2}{\z@}%
  {1.0ex plus 0.8ex minus 0.2ex}{0.3ex plus .2ex minus 0ex}%
  {\normalfont\normalsize\itshape}}
\renewcommand{\subsubsection}{\@startsection{subsubsection}{3}{\parindent}%
  {0ex plus 0.1ex minus 0.1ex}{0ex}{\normalfont\normalsize\itshape}}
\renewcommand{\paragraph}{\@startsection{paragraph}{4}{2\parindent}%
  {0ex plus 0.1ex minus 0.1ex}{0ex}{\normalfont\normalsize\itshape}}
\makeatother

\begin{document}
\renewcommand{\baselinestretch}{0.820}\selectfont

\title{\LARGE Trading Transparency for Stability:
Friction-Cone-Constrained Bilateral Teleoperation
for Opening Doors of Unsecured Furniture}

\author{\IEEEauthorblockN{Cheng Zhou, Jing Li}%
\thanks{Cheng Zhou and Jing Li are with Tencent Robotics X, Shenzhen,
China. E-mail: chowchzhou@tencent.com (Cheng Zhou) and
li\_jing3616@163.com (Jing Li).}}

\maketitle

\begin{abstract}
Opening the door of unsecured furniture loads the frictional
contacts at its supports; too large a pull slides or tips the object.
A friction-cone constraint on the commanded wrench removes this
failure mode but costs haptic transparency. We quantify the
trade-off: an admissible wrench set built from contact friction cones
and support-polygon tipping margins is shrunk for robustness to
bounded mass and friction errors, and any command in the shrunken set
provably leaves the furniture static
(Prop.~\ref{prop:static}). We call this map the \emph{robust
friction-cone projection} (RFCP). RFCP and its haptic guard
admit a bounded force transmission error
(Prop.~\ref{prop:transparency}): transparency degrades only within
a band near the constraint boundary, whose width is set by one
scalar, the stability factor $\eta$. A working-condition
controller handles the door: the operator commands a scalar swing
amplitude, the slave orientation follows hinge kinematics, and the
grasp wrench is regulated per phase. A pre-task certificate decides
whether a door can be opened without moving the furniture.
Experiments on three cabinets and two supports with five
operators ($100$ trials) measure the stability--transparency
trade-off and ablate the guard, RFCP and the position
baseline. The mechanism is not door-specific: it applies to any body
reacted by frictional contacts.
\end{abstract}

\begin{IEEEkeywords}
Bilateral teleoperation, haptic transparency, friction cone,
multi-contact stability, shared control.
\end{IEEEkeywords}

\section{INTRODUCTION}
\begin{figure}[!t]
\centering
\includegraphics[width=0.90\columnwidth]{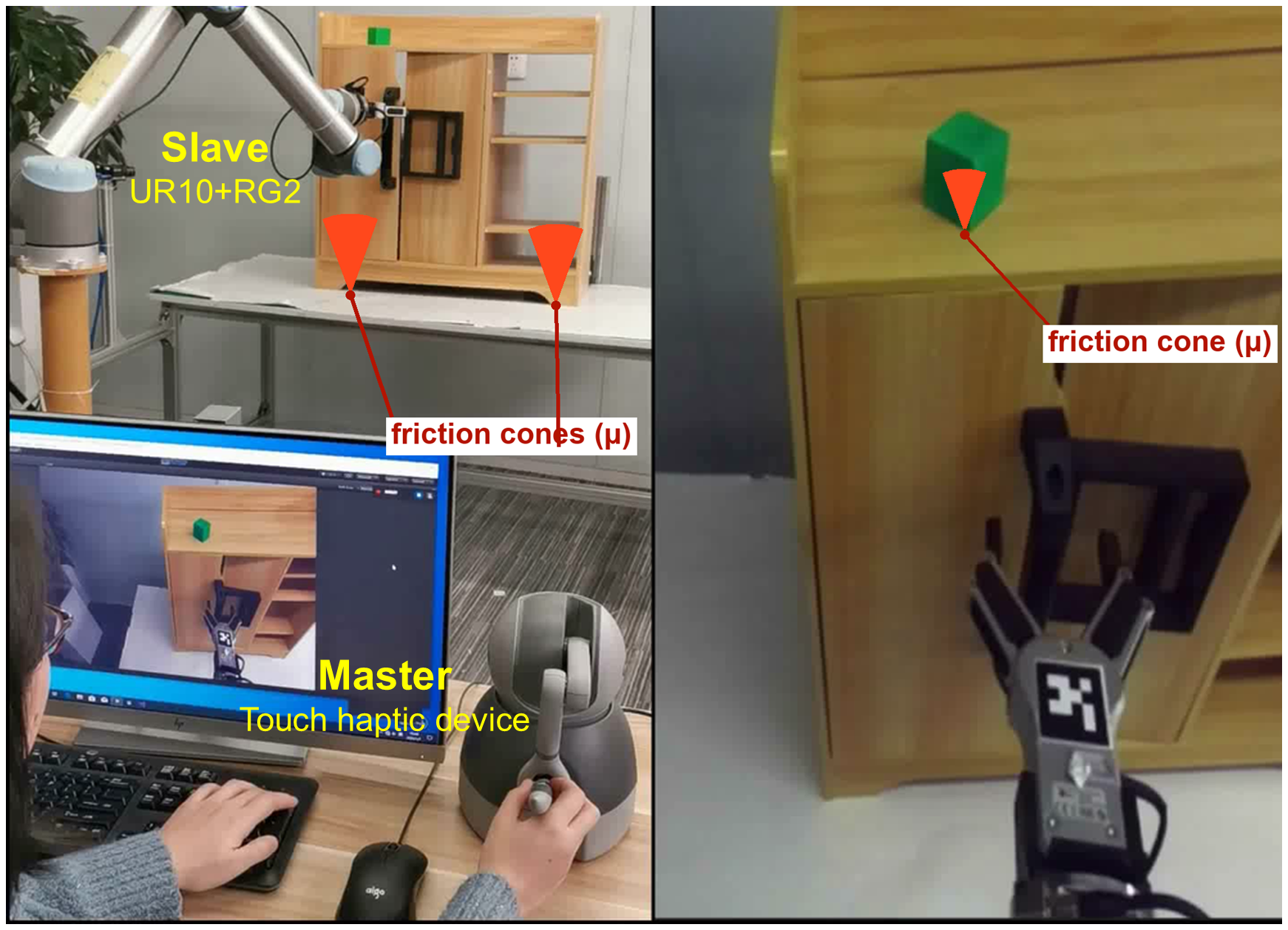}
\caption{Overview of the system. Left: the operator drives the Touch
haptic device (master) while the UR10$+$RG2 (slave) opens the door of a
cabinet resting \emph{unsecured} on the table. Right: grasp close-up
with the cone at the load--cabinet contact. Symbols: the red wedges
mark the friction cones at the cabinet--support contacts, of half-angle
$\arctan\mu_i$~\eqref{eq:ellipse}; the green cube is the tracked body
marker whose displacement measures furniture motion.}
\label{fig:teaser}
\end{figure}

\IEEEPARstart{H}{aptic} teleoperation remains the most reliable
control paradigm for contact-rich manipulation in unstructured
environments~\cite{majeed2025}. A common assumption, rarely stated
and often violated, is that the manipulated object is fixed to the world.
Opening the door of a cabinet that merely rests on the floor or a
table invalidates it: the wrench applied at the handle is reacted by
Coulomb contacts between the furniture and its support, and if that
wrench leaves the contact wrench cone the whole object slides or
tips~\cite{zhou2016,caron2017} (Fig.~\ref{fig:teaser}). The door
opens, the scene is destroyed, and, with the operator looking through
the robot's camera, the motion may go unnoticed. Two facts make this more than a
corner case: for a tall cabinet the tipping bound is \emph{tighter}
than the sliding one, not the one operators guard against, and
latch breakaway needs a force spike of unknown magnitude. Prior work
addressed stability under
delay~\cite{tho2023,heck2018,ryu2004,black2025},
transparency~\cite{majeed2025,zakerimanesh2021,hejrati2023,piccinelli2025},
shared control and haptic
arbitration~\cite{bustamante2024,fu2025} and
autonomous door
opening~\cite{wang2025};
but the stability of the \emph{manipulated object} has not been
encoded into the teleoperation channel, nor its transparency cost
quantified.

This paper treats the problem as a trade-off rather than as a
feature list. Concretely:

\begin{enumerate}
\item \emph{Staticity (Prop.~\ref{prop:static}).} From the support
wrench cone (friction, spin and tipping, Lemma~\ref{lem:cwc}) we build
a robust admissible set that tolerates bounded mass and friction
error; every wrench in it admits a feasible contact reaction.
\item \emph{Transparency cost and its floor
(Prop.~\ref{prop:transparency}, Thm.~\ref{thm:converse}).} The
felt-wrench error is zero inside the robust set and grows only in a
boundary band whose width is the margin given up for robustness. No
memoryless staticity-preserving map does better, since friction is
unobservable until the furniture moves; RFCP attains this floor
(Cor.~\ref{cor:tight}).
\item \emph{Door controller and measured trade-off.} A five-state
controller maps master position to a scalar swing amplitude and
solves orientation from hinge kinematics (Lemma~\ref{lem:track});
100 trials (five operators, three cabinets, two supports) ablate
guard, projection and baseline.
\item \emph{Force reflection as shared control.} We give the displayed
impedance (Lemma~\ref{lem:disp}) and passivity
(Prop.~\ref{prop:passivity}), and show the projection is a Lipschitz
blend of operator and feasible wrench (Lemma~\ref{lem:arb}): shared
control whose partner is contact mechanics, not a policy.
\end{enumerate}

We claim no new friction model (the mechanics is
classical~\cite{zhou2016,caron2017}), but the robust projection
inside the \emph{teleoperation} loop, its certificate, its tight
converse and an account of what the guarantee costs in transparency.

\section{RELATED WORK}

\emph{Bilateral teleoperation.} Transparency is characterised by
two-port criteria and ideal kinesthetic
coupling~\cite{majeed2025,zakerimanesh2021}; four-channel
architectures~\cite{majeed2025} widen the transparent class and
distributed passivity scales it to coordinated
slaves~\cite{yang2024}; stability under delay comes from scattering/wave
variables~\cite{tho2023,heck2018} and time-domain
passivity~\cite{ryu2004}, flown at 800\,ms
delay~\cite{panzirsch2022}, its sampled-data bound
sharpened~\cite{roberts2024} and extended to constrained
leaders~\cite{kastritsi2025}; recent work bounds the rendered force by a
small $L_2$ gain~\cite{zhang2024}, keeps the human explicitly in the
loop~\cite{black2025}, enforces synchronisation under varying
delay~\cite{zakerimanesh2021} and makes a linear MPC
passive with a linearised tank~\cite{piccinelli2025}. These guarantee
the \emph{channel}: the energy exchanged with the operator, or
operator--slave agreement. None constrains the \emph{object}, i.e.\
whether the body being manipulated stays where it was. That is the
constraint we add, and the one that makes the problem a trade-off
rather than a free gain. Proposition~\ref{prop:transparency} takes a
transmission error in the spirit of~\cite{majeed2025} but
specialises it to the projection, so the bound is \emph{a priori};
Lemma~\ref{lem:disp} recasts the classical limit as a displayed
impedance. The slave uses impedance--admittance
control~\cite{keemink2018} and operational-space force
control~\cite{herzog2016}; coupled stability against a stiff
environment~\cite{hejrati2023} and adaptive impedance in contact
tasks~\cite{michel2021} are classical. Friction
cones decide whether a compliantly supported part moves under a
manipulator load~\cite{hogan2020,teng2025}, the principle we apply to
furniture; the same cone bounds the contact wrench in dynamic
nonprehensile manipulation~\cite{zhou2022iros,zhou2023iros}, where it
is a planning constraint rather than a filter.

\emph{Shared control.} Authority is either \emph{blended}, the
executed command being a convex combination
$(1-\alpha)\bv{u}_h+\alpha \bv{u}_a$~\cite{bustamante2024}, recently
a control-barrier-function
certificate~\cite{uzun2025}, or \emph{arbitrated over
a goal set}, the weight coming from intent, confidence or
hindsight~\cite{fu2025,gulecyuz2025}; haptic instantiations render
virtual fixtures encoding \emph{geometric}
constraints~\cite{tian2024}, learned from
demonstration~\cite{muhlbauer2024}, reacting to incipient
contact~\cite{wangsafety2025} and checking feasibility before a
command is executed~\cite{bustamante2024}. In every case the
partner is a policy, a goal set or a model of the human, and except
where contact is re-parameterised
explicitly~\cite{duan2024} or the wrench is confined to a cone
by construction~\cite{teng2025}, it is blind to whether the
object touched is itself equilibrated. Ours is of the first
kind: by Lemma~\ref{lem:arb} the projection is exactly such a
blend, its partner being the contact mechanics of the object,
feasible by construction, continuous, its cost bounded a priori
(Prop.~\ref{prop:transparency}) and irreducible
(Thm.~\ref{thm:converse}), a floor that a learned arbitrator
inherits rather than removes.

\emph{Contact stability and door opening.} Closed-form wrench cones
and ZMP support areas~\cite{caron2017} make multi-contact equilibrium
testable in closed form; we adopt that machinery not for analysis but
as a real-time command filter, which robust formulations
maximise as a friction stability margin~\cite{shirai2024}, while
data-driven force--motion
models~\cite{zhou2016,hogan2020} predict whether a pushed body will
slide and are identified, as ours is, from slow force ramps. Door
opening~\cite{wang2025}, tactile manipulation of
articulated objects~\cite{zhao2024}, compliant doorway
traversal~\cite{thamrongaphichartkul2024} and force-space skill
formulations~\cite{fang2025} all constrain the
\emph{robot--object} contact: how to grasp the handle, when the latch
yields, which trajectory admits a feasible wrench. We constrain the
\emph{object--environment} one, whether the furniture stays
put, which none of the above writes, and we write it \emph{inside}
a bilateral loop, where the constraint is felt rather than planned:
the operator is told through force that the command was admissible,
and pays for that guarantee in transparency
(Thm.~\ref{thm:converse}).

\section{PROBLEM STATEMENT}
\label{sec:problem}

The slave is a 6-DoF arm with a parallel gripper and a wrist
6-axis force/torque sensor; the master is a 6-DoF haptic device with
3-DoF force rendering. The furniture (mass $m$) rests on a horizontal
support through $n$ contacts $\bv{c}_1,\dots,\bv{c}_n$ with Coulomb
coefficients $\mu_i$; the robot grasps the door handle at $\bv{H}$ and
applies a wrench $\bv{W}=(\bv{F}_h,\bv{M}_h)$. The door rotates about hinge axis
$(\bv{o},\hat{\bv{a}})$ by angle $\theta$, $\theta=0$ closed. The camera provides handle pose, hinge geometry and a body
marker. Estimated quantities carry a hat; commanded wrenches before
and after projection are $\bv{W}_d$ and $\bv{W}^{*}$.

\begin{figure}[t]
\centering
\resizebox{0.82\columnwidth}{!}{%
\begin{tikzpicture}[>=Stealth]
\tikzset{
  blk/.style={draw,rounded corners=2pt,line width=0.6pt,align=center,
              inner sep=3pt,font=\scriptsize},
  op/.style={blk,fill=gray!12,draw=gray!70!black},
  mst/.style={blk,fill=red!8,draw=red!65!black},
  ctl/.style={blk,fill=blue!7,draw=blue!70!black},
  chip/.style={draw,rounded corners=1.5pt,line width=0.4pt,fill=white,
               font=\tiny,inner sep=2pt,align=center},
  ar/.style={->,thick,draw=gray!75!black},
  fb/.style={->,thick,draw=red!65!black},
  lab/.style={font=\tiny,inner sep=1.5pt}
}
\node[op] (opr) at (4.6,5.80) {Operator};
\node[mst,minimum width=5.2cm,minimum height=0.95cm] (mast) at (4.6,4.65)
  {\textbf{Master}: PC2$\cdot$Unity\\[1pt]
   Phantom device, renders $\bv{f}_m$~\eqref{eq:fmaster}};
\node[ctl,minimum width=5.6cm,minimum height=2.20cm] (ctl) at (4.6,2.72) {};
\node[ctl,minimum width=5.2cm,minimum height=0.95cm] (slv) at (4.6,0.65)
  {\textbf{Slave}: UR10 $+$ RG2\\[1pt]
   wrist F/T returns $\bv{f}_e$};
\node[font=\scriptsize] at (4.6,3.48) {\textbf{Controller}: PC1$\cdot$ROS};
\node[chip] (th) at (2.9,2.72) {$\theta_d$~\eqref{eq:amplitude}};
\node[chip] (hp) at (4.6,2.72) {$\mat{T}_h^d$~\eqref{eq:handlepose}};
\node[chip] (wd) at (6.3,2.72) {$\bv{W}_d$};
\draw[ar,line width=0.5pt] (th)--(hp);
\draw[ar,line width=0.5pt] (hp)--(wd);
\node[chip,minimum width=4.4cm] (pr) at (4.6,2.06)
  {$\Pi_{\hat\cF(\eta)}\!:\ \bv{W}_d\mapsto \bv{W}^{*}$~\eqref{eq:proj}};
\draw[ar,line width=0.5pt] (wd.south)--(wd.south|-pr.north);
\draw[ar] (opr)--(mast) node[midway,right,lab] {hand pose $\bv{p}_m$};
\draw[ar] (mast)--(ctl) node[pos=0.42,right,lab] {$\bv{p}_m$};
\draw[ar] (pr.south)--(slv.north) node[midway,right,lab] {$\bv{W}^{*}$};
\draw[fb] (slv.west)--(0.85,0.65)|-(mast.west);
\node[lab,rotate=90] at (0.62,2.70) {measured wrench $\bv{f}_e$};
\node[chip,draw=gray!60!black,fill=gray!8] (env) at (7.75,0.65)
  {\rotatebox{90}{cabinet $+$ contacts}};
\draw[ar,line width=0.5pt,dashed] (slv.east)--(env.west);
\end{tikzpicture}}
\caption{Block diagram of the implemented loop. \emph{Master}
(red): PC2/Unity drives the Phantom and renders
$\bv{f}_m$~\eqref{eq:fmaster}. \emph{Controller} (blue): PC1/ROS maps the
master to $\theta_d$~\eqref{eq:amplitude},
solves the handle pose~\eqref{eq:handlepose} and projects onto
$\hat\cF(\eta)$~\eqref{eq:proj}. \emph{Slave} (blue): UR10$+$RG2
executes $\bv{W}^{*}$; the wrist F/T returns $\bv{f}_e$.
Symbols: $\bv{p}_m$ measured master hand pose; $\theta_d$ commanded
door-angle trajectory; $\mat{T}_h^d$ desired handle pose; $\bv{W}_d$
and $\bv{W}^{*}$ the wrench before and after
$\Pi_{\hat\cF(\eta)}$, the projection onto the estimated admissible
set; the red feedback path carries the measured wrench back to the
master; the dashed link is the slave acting on the cabinet.}
\label{fig:archraw}
\end{figure}
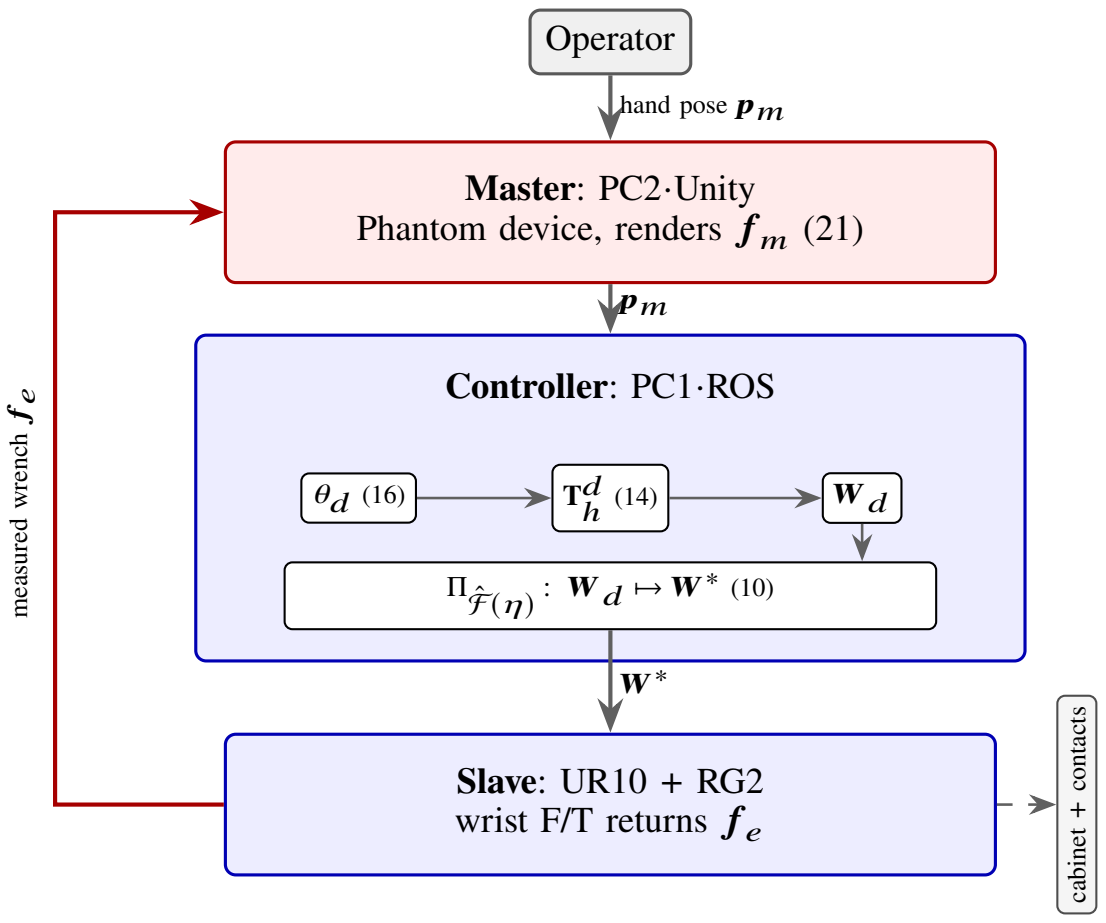

\begin{assumption}[Quasi-static interaction]
\label{as:qs}
During grasp, breakaway and swing the furniture acceleration is
negligible relative to $g$: inertial contributions are bounded by a
known $\varepsilon_{\mathrm{dyn}}$ in the wrench balance.
\end{assumption}

\begin{assumption}[Bounded parameter error]
\label{as:param}
The estimates satisfy $|\hat\mu-\mu|\le\delta_\mu$,
$|\hat m-m|\le\delta_m$ (Sec.~\ref{sec:id}), and the wrench noise is
bounded, $\|\bv{n}_W\|\le\delta_W$.
\end{assumption}

\emph{Problem.} Design (i) a command map $\bv{W}_d\mapsto \bv{W}^{*}$ keeping
the furniture static under Assumptions~\ref{as:qs}--\ref{as:param},
and (ii) a force-feedback law $\bv{f}_m$ that is stable and whose deviation
from ideal transparency is bounded and known \emph{a priori} as a
function of the stability margin. Section~\ref{sec:friction} solves
(i), Section~\ref{sec:transparency} solves (ii), Section~\ref{sec:wc}
gives the controller producing $\bv{W}_d$; Fig.~\ref{fig:archraw} shows the
implemented loop.

\section{FRICTION-CONE CONSTRAINT AND STATICITY GUARANTEE}
\label{sec:friction}

\subsection{Contact Model and Admissible Wrench Set}

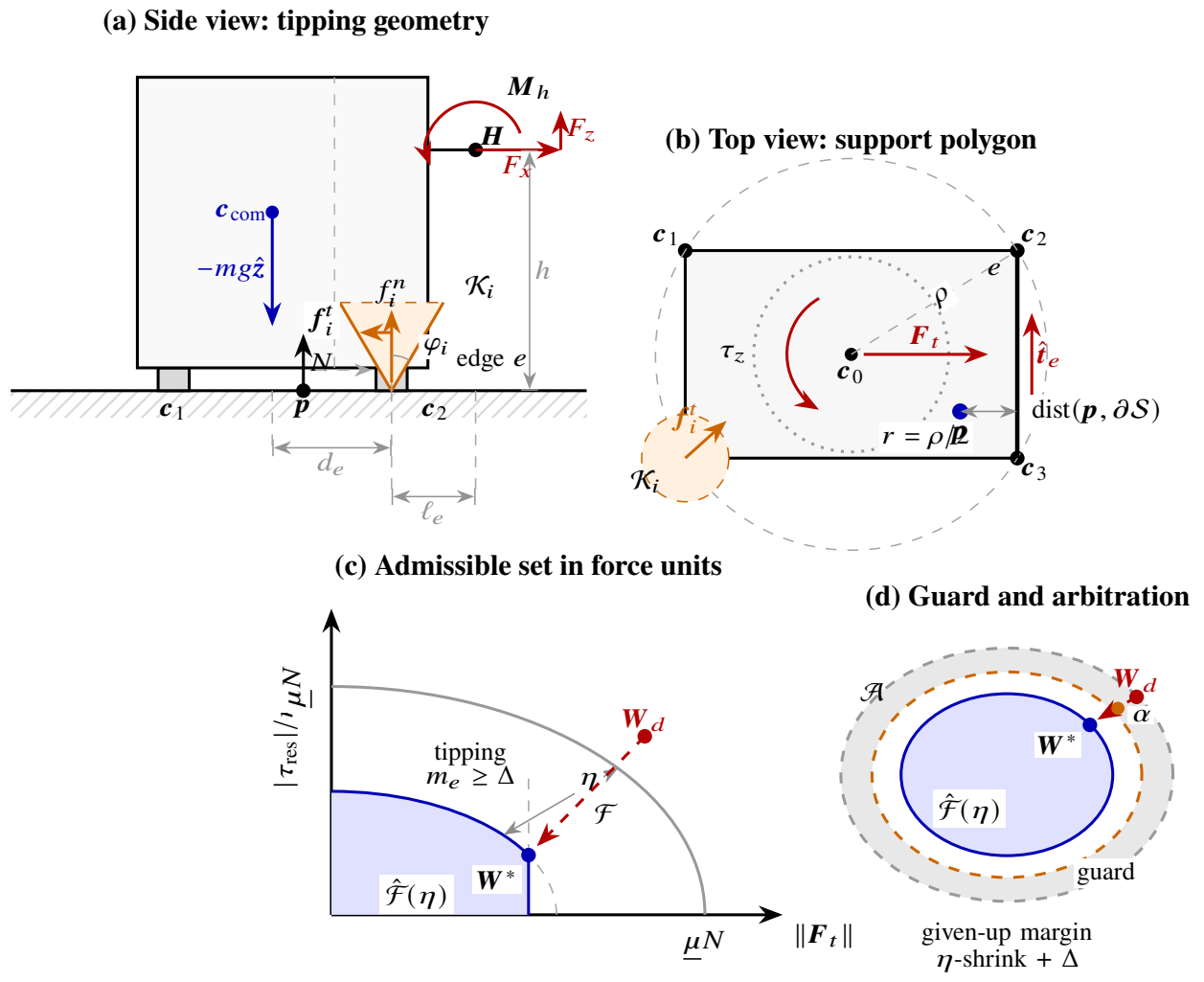
\begin{figure}[t]
\centering
\resizebox{0.93\columnwidth}{!}{%
\begin{tikzpicture}[>=Stealth]
\tikzset{
  dim/.style={|<->|,thin,gray!85},
  flab/.style={font=\scriptsize,inner sep=1.2pt},
  farc/.style={->,thick,red!70!black},
  fgrav/.style={->,thick,blue!70!black},
  ffoot/.style={->,thick,orange!80!black}
}

\begin{scope}
\fill[pattern=north east lines,pattern color=gray!30] (-0.70,-0.26) rectangle (4.85,0);
\draw[thick] (-0.70,0)--(4.85,0);
\fill[gray!30,draw=black,thick] (0.72,0) rectangle (1.02,0.22);
\fill[gray!30,draw=black,thick] (2.82,0) rectangle (3.12,0.22);
\fill[gray!5,draw=black,thick] (0.52,0.22) rectangle (3.32,3.02);
\draw[dashed,gray!60] (2.42,0.22)--(2.42,3.02);
\draw[thick] (3.32,2.32)--(3.78,2.32);
\filldraw[black] (3.78,2.32) circle (1.8pt) node[above right,flab] {$\bv{H}$};
\draw[farc] (3.78,2.32)--(4.60,2.32) node[midway,below,flab] {$F_x$};
\draw[farc] (4.60,2.32)--(4.60,2.70) node[midway,right,flab] {$F_z$};
\draw[farc] (4.21,2.48) arc (20:200:0.46);
\node[flab] at (4.30,2.92) {$\bv{M}_h$};
\filldraw[blue!70!black] (1.82,1.72) circle (1.6pt) node[left,flab] {$\bv{c}_{\mathrm{com}}$};
\draw[fgrav] (1.82,1.72)--(1.82,0.62) node[midway,left,flab] {$-mg\hat{\bv{z}}$};
\fill[orange!12] (2.97,0)--(2.47,0.85)--(3.47,0.85)--cycle;
\draw[orange!80!black,thick] (2.97,0)--(2.47,0.85) (2.97,0)--(3.47,0.85);
\draw[orange!80!black,dashed] (2.47,0.85)--(3.47,0.85);
\draw[dashed,gray!70] (2.97,0)--(2.97,0.85);
\draw[ffoot] (2.97,0.08)--(2.97,0.78);
\node[flab] at (2.97,0.94) {$f_i^{n}$};
\draw[ffoot] (2.97,0.56)--(2.66,0.56);
\node[flab,anchor=east,fill=white,inner sep=1pt] at (2.48,0.68) {$\bv{f}_i^{t}$};
\draw[thin,gray!80] (2.97,0.34) arc (90:59.5:0.34);
\node[flab,anchor=west,fill=white,inner sep=1pt] at (3.22,0.42) {$\varphi_i$};
\node[flab,anchor=west] at (3.62,1.00) {$\cK_i$};
\node[flab,anchor=west] at (3.55,0.30) {edge $e$};
\filldraw[black] (2.12,0) circle (1.8pt) node[below,flab,yshift=-1pt] {$\bv{p}$};
\draw[->,thick] (2.12,0)--(2.12,0.55) node[midway,right,flab] {$N$};
\draw[->,dashed,thin,gray!70] (2.24,0.20)--(2.78,0.20);
\draw[dim] (4.30,0)--(4.30,2.32) node[midway,right,flab] {$h$};
\draw[dim] (1.82,-0.55)--(2.97,-0.55) node[midway,below,flab] {$d_e$};
\draw[dim] (2.97,-1.02)--(3.78,-1.02) node[midway,below,flab] {$\ell_e$};
\draw[dashed,thin,gray!70] (1.82,0)--(1.82,-0.55);
\draw[dashed,thin,gray!70] (2.97,0)--(2.97,-1.02);
\draw[dashed,thin,gray!70] (3.78,0)--(3.78,-1.02);
\node[flab] at (0.87,-0.20) {$\bv{c}_1$};
\node[flab,anchor=west] at (3.22,-0.20) {$\bv{c}_2$};
\node[font=\footnotesize\bfseries] at (2.05,3.55) {(a) Side view: tipping geometry};
\end{scope}

\begin{scope}[shift={(7.40,0.35)}]
\fill[gray!6,draw=black,thick] (-1.60,1.00)--(1.60,1.00)--(1.60,-1.00)--(-1.60,-1.00)--cycle;
\filldraw[black] (-1.60,1.00) circle (1.8pt) node[above left,flab] {$\bv{c}_1$};
\filldraw[black] (1.60,1.00) circle (1.8pt) node[above right,flab] {$\bv{c}_2$};
\filldraw[black] (1.60,-1.00) circle (1.8pt) node[below right,flab] {$\bv{c}_3$};
\filldraw[black] (-1.60,-1.00) circle (1.8pt) node[below left,flab] {$\bv{c}_4$};
\filldraw[black] (0,0) circle (1.6pt) node[below,flab,yshift=-2pt] {$\bv{c}_0$};
\draw[dashed,gray!70] (0,0) circle (1.887);
\draw[dashed,gray!70] (0,0)--(1.60,1.00);
\node[flab,rotate=32,fill=white] at (0.88,0.55) {$\rho$};
\draw[dotted,thick,gray!80] (0,0) circle (0.94);
\node[flab,fill=white,inner sep=0.8pt] at (0.72,-0.80) {$r=\rho/2$};
\draw[very thick] (1.60,-1.00)--(1.60,1.00);
\draw[->,thick,red!70!black] (1.74,-0.40)--(1.74,0.40) node[midway,right,flab] {$\hat{\bv{t}}_e$};
\node[flab,anchor=east] at (1.50,0.82) {$e$};
\filldraw[blue!70!black] (1.05,-0.55) circle (2pt);
\node[flab] at (1.05,-0.78) {$\bv{p}$};
\draw[<->,thin,gray!90] (1.05,-0.55)--(1.60,-0.55);
\node[flab,anchor=west,fill=white,inner sep=1pt] at (1.70,-0.55) {$\dist(\bv{p},\partial\cS)$};
\draw[farc] (0.12,0)--(1.32,0) node[midway,above,flab] {$\bv{F}_t$};
\draw[farc] (-0.31,0.54) arc (120:240:0.62);
\node[flab,anchor=east] at (-0.95,0.00) {$\tau_z$};
\draw[dashed,orange!80!black,fill=orange!12] (-1.60,-1.00) circle (0.42);
\draw[ffoot] (-1.60,-1.00)--(-1.20,-0.64) node[midway,above left,flab] {$\bv{f}_i^{t}$};
\node[flab] at (-1.98,-1.22) {$\cK_i$};
\node[font=\footnotesize\bfseries] at (0.00,2.05) {(b) Top view: support polygon};
\end{scope}

\begin{scope}[shift={(2.39,-5.05)}]
\draw[->,thick] (0,0)--(4.35,0);
\node[flab,anchor=north west] at (4.40,-0.04) {$\|\bv{F}_t\|$};
\draw[->,thick] (0,0)--(0,2.95);
\node[flab,rotate=90,fill=white] at (-0.44,1.62) {$|\tau_{\mathrm{res}}|/r$};
\draw[thick,gray!80] (3.60,0) arc (0:90:3.60 and 2.20);
\node[flab,fill=white] at (2.62,0.98) {$\cF$};
\node[flab] at (3.60,-0.30) {$\underline\mu N$};
\node[flab,rotate=90,fill=white] at (-0.32,2.20) {$\underline\mu N$};
\begin{scope}
  \clip (0,0) rectangle (1.90,3.00);
  \fill[blue!12] (0,0)--(2.17,0) arc (0:90:2.17 and 1.19)--cycle;
  \draw[blue!70!black,thick] (2.17,0) arc (0:90:2.17 and 1.19);
\end{scope}
\draw[dashed,thin,gray!70] (2.17,0) arc (0:28.8:2.17 and 1.19);
\draw[blue!70!black,thick] (1.90,0)--(1.90,0.575);
\draw[dashed,gray!70] (1.90,0.575)--(1.90,1.40);
\node[flab,align=center,fill=white,inner sep=1pt] at (1.34,1.44) {tipping\\ $m_e\ge\Delta$};
\node[flab,fill=white] at (0.82,0.20) {$\hat\cF(\eta)$};
\draw[<->,thin,gray!90] (2.76,1.41)--(1.66,0.77);
\node[flab,fill=white] at (2.48,1.28) {$\eta$};
\filldraw[red!70!black] (3.02,1.72) circle (1.8pt) node[above,flab] {$\bv{W}_d$};
\filldraw[blue!70!black] (1.90,0.575) circle (1.8pt);
\node[flab,anchor=north east,fill=white] at (1.84,0.48) {$\bv{W}^{*}$};
\draw[dashed,->,thick,red!70!black] (3.02,1.72)--(1.98,0.66);
\node[font=\footnotesize\bfseries] at (1.90,3.35) {(c) Admissible set in force units};
\end{scope}

\begin{scope}[shift={(7.30,-5.15)}]
\fill[gray!18] (1.60,1.45) ellipse (1.60 and 1.22);
\fill[white] (1.60,1.45) ellipse (1.30 and 0.99);
\fill[blue!12] (1.60,1.45) ellipse (1.02 and 0.78);
\draw[blue!70!black,thick] (1.60,1.45) ellipse (1.02 and 0.78);
\node[flab,fill=white,inner sep=1pt] at (1.24,1.12) {$\hat\cF(\eta)$};
\draw[thick,gray!80,dashed] (1.60,1.45) ellipse (1.60 and 1.22);
\draw[orange!80!black,thick,dashed] (1.60,1.45) ellipse (1.30 and 0.99);
\node[flab,anchor=west,fill=white,inner sep=0.8pt] at (2.24,0.50) {guard};
\filldraw[red!70!black] (2.85,2.20) circle (1.8pt) node[above,flab] {$\bv{W}_d$};
\draw[dashed,->,thick,red!70!black] (2.85,2.20)--(2.47,1.98);
\filldraw[blue!70!black] (2.40,1.93) circle (1.8pt);
\node[flab,anchor=east,fill=white,inner sep=0.8pt] at (2.32,1.74) {$\bv{W}^{*}$};
\filldraw[orange!80!black] (2.67,2.09) circle (1.6pt);
\node[flab,anchor=west,fill=white,inner sep=0.8pt] at (2.76,2.02) {$\alpha$};
\node[flab] at (0.30,2.25) {$\cA$};
\node[flab,align=center] at (1.60,-0.08) {given-up margin};
\node[flab,align=center] at (1.60,-0.34) {$\eta$-shrink $+\ \Delta$};
\node[font=\footnotesize\bfseries] at (1.80,3.15) {(d) Guard and arbitration};
\end{scope}

\end{tikzpicture}}
\caption{Friction-cone geometry and notation. (a)~Side view: the pull
$\bv{F}_h$ and free moment $\bv{M}_h$ at $\bv{H}$, gravity at
$\bv{c}_{\mathrm{com}}$; the support carries $N=\underline m g-F_z$ at
the centre of pressure $\bv{p}$, which recedes towards the leading
edge $e$ (levers $h$, $d_e$) \eqref{eq:tipping}.
(b)~Top view: support polygon $\cS$, circumradius $\rho$, lever
$r=\rho/2$; $\dist(\bv{p},\partial\cS)$ is the tipping margin. (c)~Admissible set in force units: the ellipse
\eqref{eq:ellipse} cut by tipping, shrunk by $\eta$ and offset by
$\Delta$ into $\hat\cF(\eta)$ \eqref{eq:robust}, containing
$\bv{W}^{*}=\Pi_{\hat\cF(\eta)}(\bv{W}_d)$ \eqref{eq:proj}.
(d)~Arbitration in one plane: the grey ring between $\cA$ and
$\hat\cF(\eta)$ is the margin given up, its width the unavoidable term
of \eqref{eq:converse}; the dashed curve is where the guard
\eqref{eq:guard} pushes back and $\alpha$ \eqref{eq:arbw} how far
along $\bv{W}_d-\bv{W}^{*}$ the command is honoured.}
\label{fig:cone}
\end{figure}

Fig.~\ref{fig:cone} fixes the notation. Wrenches are written about the
centroid $\bv{c}_0$ of the support polygon
$\cS=\operatorname{conv}\{\bv{c}_1,\dots,\bv{c}_n\}$; the robot applies
a force $\bv{F}_h$ and a \emph{free moment} $\bv{M}_h$ (a gripper
twisting the handle) at $\bv{H}$. The applied, gravity and contact
wrenches are
\begin{align}
\bv{W} &= \bigl(\bv{F}_h,\ (\bv{H}-\bv{c}_0)\times \bv{F}_h + \bv{M}_h\bigr),
\label{eq:W}\\
\bv{W}_g &= \bigl(-m g\,\hat{\bv{z}},\ (\bv{c}_{\mathrm{com}}-\bv{c}_0)\times(-m g\,\hat{\bv{z}})\bigr),
\nonumber\\
\bv{w} &= \sum_{i=1}^{n}\bigl(\bv{f}_i,\ (\bv{c}_i-\bv{c}_0)\times \bv{f}_i\bigr),
\nonumber
\end{align}
with $\bv{f}_i$ in the Coulomb cone
$\cK_i=\{f^{n}\hat{\bv{z}}+\bv{f}^{t}:\
\|\bv{f}^{t}\|\le\mu_i f^{n},\ f^{n}\ge0\}$ at $\bv{c}_i$
(half-angle $\varphi_i=\arctan\mu_i$).

The transport term in~\eqref{eq:W} has a lever of the order of the
furniture height, gravity only $d_e$. Under Assumption~\ref{as:qs} a
command is admissible iff the reaction it demands lies in the
\emph{support wrench cone}~\cite{caron2017} ($\oplus$: Minkowski sum):
\begin{align}
\bv{w}+\bv{W}+\bv{W}_g &= 0 ,
\label{eq:balance}\\
\cW &= \cC_1\oplus\cdots\oplus\cC_n ,%
\cF = \bigl\{\bv{W}:\ -(\bv{W}+\bv{W}_g)\in\cW\bigr\} ,
\label{eq:admissible}
\end{align}
$\cC_i$ the contact wrench cone $\{(\bv{f},(\bv{c}_i-\bv{c}_0)\times\bv{f}):\
\bv{f}\in\cK_i\}$. $\cW$ couples force and moment: translation, spin
and tipping draw on one friction reserve, which a scalar force limit
misses.

\begin{lemma}[Support wrench cone lemma: friction, spin, tipping]
\label{lem:cwc}
Let $\bv{w}=(\bv{f},\tau)\in\cW$, let
$p=(\sum_i f_i^{n}\bv{c}_i)/(\sum_i f_i^{n})$ be the centre of pressure
(CoP), $\rho=\max_i\|\bv{c}_i-\bv{c}_0\|$ the footprint circumradius,
$\bar\mu=\max_i\mu_i$ and $\underline\mu=\min_i\mu_i$.
Then, necessarily,
\begin{enumerate}
\item[(i)] $f_z\ge0$ and $\|\bv{f}_t\|\le\bar\mu f_z$
\quad(\emph{translation capacity});
\item[(ii)] $\bigl|\tau_z-(\bv{p}\times \bv{f}_t)_z\bigr|\le\bar\mu\,\bar r\,f_z$
with $\bar r=\tfrac23\rho$
\quad(\emph{spin capacity});
\item[(iii)] $\bv{p}\in\cS$ \quad(\emph{no tipping}).
\end{enumerate}
Conversely, $\bv{w}\in\cW$ whenever $\bv{p}\in\cS$ and, with
$r=\tfrac12\rho$,
\begin{equation}
\Bigl(\frac{\|\bv{f}_t\|}{\underline\mu f_z}\Bigr)^{2}
\;+\;
\Bigl(\frac{\bigl|\tau_z-(\bv{p}\times \bv{f}_t)_z\bigr|}
{\underline\mu\,r\,f_z}\Bigr)^{2}
\;\le\; 1 .
\label{eq:ellipse}
\end{equation}
\end{lemma}

\begin{IEEEproof}
\emph{Sketch.} (i) is the triangle inequality; (ii) splits $\tau_z$
about $\bv{p}$ and bounds the lever sum by $\frac23\rho f_z$; (iii)
defines the CoP. For sufficiency, distribute $f_z$ with centre
$\bv{p}\in\cS$ and set
$\bv{f}_i^t=f_i^{n}(\bv{f}_t/f_z+\lambda\,\hat{\bv{z}}\times(\bv{c}_i-\bv{p}))$;
the circulation term has no resultant, and~\eqref{eq:ellipse} keeps
every $\bv{f}_i\in\cK_i$.
\end{IEEEproof}

Lemma~\ref{lem:cwc} turns~\eqref{eq:admissible} into three scalar
tests run every millisecond. With $x$ along the pull, $z$ up, normal
load $N\triangleq\underline m g-F_z$ and CoP
\begin{equation}
\bv{p}(\bv{W})\;=\;\bv{c}_0+\frac{1}{N}\,\mat{Q}\Bigl[\bv{M}_h+(\bv{H}-\bv{c}_0)\times \bv{F}_h+\bv{M}_g\Bigr]_{xy}
\nonumber
\end{equation}
($\mat{Q}$: in-plane $90^{\circ}$ rotation), the tests read:

\emph{Slip and spin.} With $\bv{f}_t=-\bv{F}_t$, $f_z=N$,
\eqref{eq:ellipse} gives the cone at worst-case friction and the
\emph{residual} spin demand
\begin{align}
\|\bv{F}_t\| &\le \underline\mu\,N ,
\label{eq:slide}\\
\tau_{\mathrm{res}}(\bv{W}) &=
\Bigl[\bv{M}_h+(\bv{H}-\bv{c}_0)\times\bv{F}_h\Bigr]_z
+\bigl(\bv{p}(\bv{W})\times\bv{F}_t\bigr)_z ,
\label{eq:spin}
\end{align}
coupled to~\eqref{eq:slide} through~\eqref{eq:ellipse}.

\emph{Tipping.} Condition (iii) is an edge-moment inequality for every
edge $e$ (tangent $\hat{\bv{t}}_e$, $M_{h,e}=\bv{M}_h\cdot\hat{\bv{t}}_e$,
$\ell_e$ the offset of $\bv{H}$):
\begin{align}
m_e &= \Bigl[\,m g\,(\bv{c}_{\mathrm{com}}-\bv{c}_e)\times(-\hat{\bv{z}})
\nonumber\\
&\quad +(\bv{H}-\bv{c}_e)\times\bv{F}_h+\bv{M}_h\,\Bigr]\cdot\hat{\bv{t}}_e
\;\ge\;0 ,
\label{eq:tip}\\
F_x\,h+M_{h,e} &\le m g\,d_e+\ell_e F_z ,
\label{eq:tipping}
\end{align}
the second line being the leading edge. Lifting ($F_z>0$) helps
tipping but hurts sliding.

\emph{Robust set.} $\cF$ is monotone in $\mu$ and $m$; evaluating it
at $\underline\mu=\hat\mu-\delta_\mu$, $\underline m=\hat m-\delta_m$,
scaling capacities by $\eta$ and subtracting the budget
$\Delta\triangleq\delta_W+\varepsilon_{\mathrm{dyn}}+\delta_{\mathrm{kin}}$
(wrench noise, leaf inertia, grasp parasite) gives
\begin{equation}
\hat\cF(\eta)\;=\;\Bigl\{\bv{W}:\;
\begin{aligned}[t]
& m_e(\bv{W})\big|_{\eta\underline m}\;\ge\;\Delta
\quad \forall e\in\partial\cS,\\[1mm]
&\Bigl(\frac{\|\bv{F}_t\|}{\eta\underline\mu N-\Delta}\Bigr)^{2}
+\Bigl(\frac{|\tau_{\mathrm{res}}(\bv{W})|}
{\eta\underline\mu r N-\Delta}\Bigr)^{2}\;\le\;1
\end{aligned}
\Bigr\} ,
\label{eq:robust}
\end{equation}
with $m g\to\eta\underline m g$ in~\eqref{eq:tip}. Since the leaf
\emph{binds} above $F_{\mathrm{jam}}$ (Sec.~\ref{sec:exp}),
$\hat\cF(\eta)$ is also capped by $\|\bv{F}\|\le\bar F_{\mathrm{jam}}$.

\begin{definition}[Robust friction-cone projection (RFCP)]
\label{def:rfcp}
The door-controller wrench $\bv{W}_d$ is applied as
\begin{equation}
\bv{W}^{*} \;=\; \operatorname*{arg\,min}_{W\in\hat\cF(\eta)}
\ \bigl\|\bv{W} - \bv{W}_d\bigr\|_{\bv{\Lambda}}^{2},
\label{eq:proj}
\end{equation}
with $\bv{\Lambda}=\mathrm{diag}(\mat{I}_3,r^{-2}\mat{I}_3)$
(moments in force units). It is solved by cyclic projection onto the
disc~\eqref{eq:slide}, the tipping bound~\eqref{eq:tipping} and the
ellipse~\eqref{eq:ellipse}, convex sets, so it converges; stopping at a
$0.1\,\mathrm{N}$ increment takes three passes in $98\%$ of cycles.
\end{definition}

\begin{lemma}[RFCP is contractive and firmly non-expansive]
\label{lem:proj}
Let $\hat\cF(\eta)$ be closed and convex with $0\in\hat\cF(\eta)$,
and let $\bv{\Lambda}\succ0$. Then for all $\bv{W},\bv{W}'\in\mathbb R^6$:
\begin{enumerate}
\item[(i)] $\|\Pi(\bv{W})\|_{\bv{\Lambda}}\le\|\bv{W}\|_{\bv{\Lambda}}$ and
$\|\Pi(\bv{W})-\Pi(\bv{W}')\|_{\bv{\Lambda}}\le\|\bv{W}-\bv{W}'\|_{\bv{\Lambda}}$
\quad(\emph{contraction}, \emph{non-expansiveness});
\item[(ii)] $\|\Pi(\bv{W})-\Pi(\bv{W}')\|_{\bv{\Lambda}}^{2}
+\|(\mat{I}-\Pi)(\bv{W})-(\mat{I}-\Pi)(\bv{W}')\|_{\bv{\Lambda}}^{2}\le\|\bv{W}-\bv{W}'\|_{\bv{\Lambda}}^{2}$
\quad(\emph{firm non-expansiveness});
\item[(iii)] $\langle \Pi(\bv{W}),\,\bv{W}\rangle_{\bv{\Lambda}}\ge
\|\Pi(\bv{W})\|_{\bv{\Lambda}}^{2}\ge0$
\quad(\emph{monotonicity}, i.e.\ passivity of $\Pi$ read as a static
operator $\bv{W}_d\mapsto \bv{W}^{*}$).
\end{enumerate}
\end{lemma}

\begin{IEEEproof}
Standard for Euclidean projection onto a closed convex set containing
$0$, applied in the coordinates $\bv{\Lambda}^{1/2}\bv{W}$.
\end{IEEEproof}

The filter only \emph{shrinks} a command (i) and injects no energy
(iii), so it can sit in a passive loop (Prop.~\ref{prop:passivity}).

\begin{proposition}[Staticity]
\label{prop:static}
Let Assumptions~\ref{as:qs}--\ref{as:param} hold and let
$\bv{W}^{*}=\Pi_{\hat\cF(\eta)}(\bv{W}_d)$ be the applied wrench. Then there
exist contact reactions $\{\bv{f}_i\}$, $\bv{f}_i\in\cK_i$, and a centre of
pressure $\bv{p}\in\cS$ such that ~\eqref{eq:balance} holds with the
\emph{true} $(\mu,m)$: starting from rest, the furniture neither
slides, spins nor tips, with slack $\eta$ on all three margins.
\end{proposition}

\begin{IEEEproof}
The three margins of~\eqref{eq:robust} are monotone in $(\mu,m)$, so
with $\underline\mu\le\mu$, $\underline m\le m$, $\eta\le1$ they hold
for the true parameters; the sufficiency part of Lemma~\ref{lem:cwc}
then yields the reactions, $\Delta$ covering noise, inertia
and~\eqref{eq:kineerr}.
\end{IEEEproof}

Transients exceeding $\Delta$ (latch breakaway, re-grasp impact) lie
outside this guarantee; keeping the operator out of them is the
guard's task.

\subsection{Projection, Haptic Guard and the Stability--Transparency Knob}
The projection is invisible; to make the constraint \emph{legible}
before it binds, the operator feels a repulsive guard:
\begin{equation}
\bv{f}_{\mathrm{vf}}\;=\;-k_g\,\bigl[\,g(\bv{W}_d)\,\bigr]_{+}^{2}\,\hat{\bv{n}}_g,
\qquad
g(\bv{W}_d)=\frac{\dist\bigl(\bv{W}_d,\partial\hat\cF\bigr)}{F_{\mathrm{ref}}},
\label{eq:guard}
\end{equation}
with $\hat{\bv{n}}_g$ the outward normal in master coordinates and
$F_{\mathrm{ref}}$ a normalizing force. $\eta$ sets how early the hard
limit binds, $k_g$ how softly; only $\eta$ enters
Prop.~\ref{prop:static}.

\begin{remark}[Task feasibility certificate]
\label{rem:feas}
If the breakaway force exceeds the bounds of~\eqref{eq:robust},
\emph{no} wrench opens the door without moving the furniture. Hence
$F_{\mathrm{bw}}\le\min\{\eta\underline\mu\underline m g,\
\eta\underline m g\,d_e/h\}-\delta_W-\varepsilon_{\mathrm{dyn}}$,
checked from a force ramp, is a pre-task certificate; on C1 it gives
$26\,\mathrm{N}$ at $\eta=0.7$ against a measured
$14$--$22\,\mathrm{N}$.
\end{remark}

\subsection{The Constraint as a Shared-Control Arbitrator}
The operator decides \emph{what} to do, the slave controller supplies
orientation (Sec.~\ref{sec:wc}), and the contact model adjudicates
\emph{how much} of the command the environment can carry; in
shared-control form, \eqref{eq:proj} is itself an arbitrator.

\begin{lemma}[Lipschitz arbitration weight of RFCP]
\label{lem:arb}
Let $\bv{W}_d$ violate exactly one constraint of $\hat\cF(\eta)$. Then
\eqref{eq:proj} is a convex blend of the human command $\bv{W}_d$ and an
assistive command $\bv{W}_a$ that keeps the operator's intent but zeroes
the offending component, $\alpha$ being the authority ceded to it:
\begin{align}
\bv{W}^{*} &= (1-\alpha)\,\bv{W}_d+\alpha\,\bv{W}_a , && \alpha\in[0,1) ,
\label{eq:arb}\\
\alpha &= \frac{[\|\bv{F}_t^{d}\|-c_t]_{+}}{\|\bv{F}_t^{d}\|},&
\bv{W}_a &= \bv{W}_d-\bv{F}_t^{d}
\quad\text{(disc),} \nonumber\\[1mm]
\alpha &= \frac{[\langle\hat{\bv{n}},\bv{W}_d\rangle-b]_{+}}
{\langle\hat{\bv{n}},\bv{W}_d\rangle} , && \text{(half-space)}
\label{eq:arbw}\\[1mm]
\bv{W}_a &= \bv{W}_d-\frac{\langle\hat{\bv{n}},\bv{W}_d\rangle}
{\|\hat{\bv{n}}\|_{\bv{\Lambda}^{-1}}^{2}}\,\bv{\Lambda}^{-1}\hat{\bv{n}} ,
\nonumber
\end{align}
and: (i) $\alpha=0$ iff the constraint is inactive and
$0<\alpha<1$ puts $\bv{W}^{*}$ exactly on the active boundary, i.e.\ the
constraint takes exactly the authority the violation requires;
(ii) $\bv{W}_a$ differs from $\bv{W}_d$ only along the
\emph{offending} coordinate, the pull direction or the
$\bv{\Lambda}$-normal, so arbitration never rotates the operator's
intent; (iii) $\alpha$ is continuous and piecewise $C^{1}$ in the
active margin coordinate $z$ with $|\partial\alpha/\partial z|\le
1/\varsigma$, $\varsigma$ being the admissible value of that
coordinate ($c_t$ or $b$).
\end{lemma}

\begin{IEEEproof}
Substituting ~\eqref{eq:arbw} into ~\eqref{eq:arb} returns
$\bv{F}_t^{*}=(1-\alpha)\bv{F}_t^{d}$, hence $\|\bv{F}_t^{*}\|=c_t$, and
$\langle\hat{\bv{n}},\bv{W}^{*}\rangle=\min\{\langle\hat{\bv{n}},\bv{W}_d\rangle,b\}$, the
two cases of Lemma~\ref{lem:radial}, so ~\eqref{eq:arb} is the
projection, with $\bv{W}^{*}$ on $\partial\hat\cF(\eta)$ by
$\alpha\in[0,1)$ (i); (ii) is the definition of $\bv{W}_a$. For
(iii), $\alpha=1-\varsigma/z$ on $\{z>\varsigma\}$ has derivative
$\varsigma/z^{2}\le1/\varsigma$ and $\alpha\to0$ as
$z\to\varsigma^{+}$, so $\alpha$ extends continuously by $0$.
\end{IEEEproof}

The arbitrator is thus \emph{continuous} (a generous margin means a
gentle handover) and \emph{minimal} (it overrules the operator only
as far as the command is infeasible): $\eta$ sets how much authority
the constraint is \emph{given}, $\alpha$ how much it \emph{uses}.

\section{WORKING-CONDITION DOOR CONTROLLER}
\label{sec:wc}

\subsection{Door Kinematics: Amplitude from the Master, Orientation
Solved Onboard}
With $\mat{T}_h^{0}$ the handle pose when closed, the desired pose for a
commanded angle $\theta_d$ is the closed pose rotated about the hinge,
and the slave pose follows by inverse kinematics at $\mat{T}_h^d$, so
the operator never commands orientation. Estimating $(\hat{\bv{a}},\bv{o})$
with errors $\delta_a,\delta_o$ displaces the handle by the bound
below, with $r_H\triangleq\|\bv{r}_{o\to \bv{H}}\|$:
\begin{align}
\mat{T}_h^{d}(\theta_d) &=
\mathrm{Trans}(\bv{o})\,\mathrm{Rot}(\hat{\bv{a}},\theta_d)\,
\mathrm{Trans}(-\bv{o})\;\circ\;\mat{T}_h^{0} ,
\label{eq:handlepose}\\
\|\delta\bv{p}_H\| &\le 2\,\delta_o+\theta_d\,r_H\,\delta_a ,
\label{eq:kineerr}
\end{align}
or, through a grasp of stiffness $k_{\mathrm{st}}$, the parasitic
wrench $\delta_{\mathrm{kin}}=
k_{\mathrm{st}}(2\delta_o+\theta_{\max}r_H\delta_a)$; the margin
of~\eqref{eq:robust} is then the computable sum
$\Delta=\delta_W+\varepsilon_{\mathrm{dyn}}+\delta_{\mathrm{kin}}$.

The angle setpoint is a \emph{scalar} map of the master translation
along the mapped pull direction $\hat{\bv{e}}_m=\mat{R}_b^{\T}\hat{\bv{t}}_s$,
\begin{equation}
\theta_d \;=\;\cS\Bigl\{\Pi_{[0,\theta_{\max}]}
\bigl\{k_\theta\,\sigma\bigr\}\Bigr\},\qquad
\sigma=(\bv{p}_m-\bv{p}_{\mathrm{ref}})\cdot\hat{\bv{e}}_m ,
\label{eq:amplitude}
\end{equation}
with $\Pi$ the clamp, $\bv{p}_{\mathrm{ref}}$ captured at grasp,
$k_\theta=\theta_{\max}/\ell_m$ and $\cS$ the rate limiter enforcing
the bounds of Table~\ref{tab:params}, which \emph{makes}
Assumption~\ref{as:qs} true, so $\varepsilon_{\mathrm{dyn}}$ is
enforced rather than posited.

\subsection{Working Conditions}
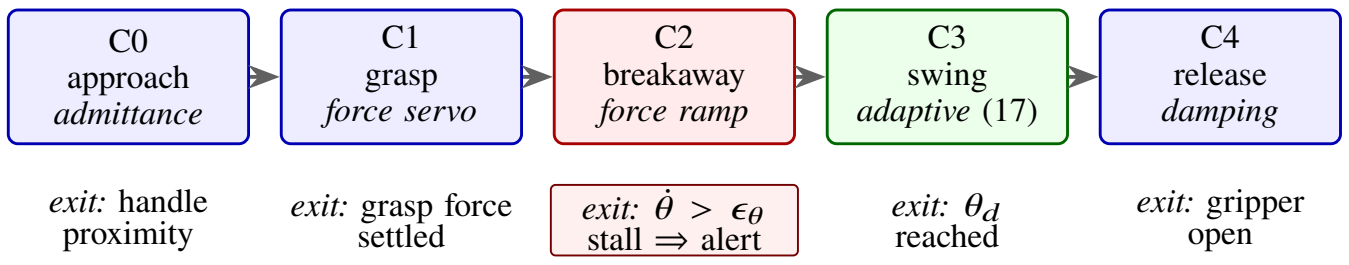
\begin{figure}[t]
\centering
\resizebox{1.0\columnwidth}{!}{%
\begin{tikzpicture}[>=Stealth]
\tikzset{
  st/.style={draw,rounded corners=2pt,line width=0.6pt,fill=blue!7,
             draw=blue!70!black,font=\scriptsize,align=center,text width=1.62cm,
             inner sep=2pt,minimum height=0.98cm},
  ex/.style={font=\scriptsize,inner sep=1pt,align=center,text width=1.74cm},
  ar/.style={->,thick,draw=gray!75!black}
}
\node[st] (c0) at (0.95,1.60) {C0\\[1pt]approach\\[1pt]\emph{admittance}};
\node[st] (c1) at (2.95,1.60) {C1\\[1pt]grasp\\[1pt]\emph{force servo}};
\node[st,fill=red!8,draw=red!65!black] (c2) at (4.95,1.60)
  {C2\\[1pt]breakaway\\[1pt]\emph{force ramp}};
\node[st,fill=green!8,draw=green!40!black] (c3) at (6.95,1.60)
  {C3\\[1pt]swing\\[1pt]\emph{adaptive~\eqref{eq:swing}}};
\node[st] (c4) at (8.95,1.60) {C4\\[1pt]release\\[1pt]\emph{damping}};
\foreach \a/\b in {c0/c1,c1/c2,c2/c3,c3/c4}{\draw[ar] (\a)--(\b);}
\node[ex] at (0.95,0.55) {\emph{exit:} handle\\proximity};
\node[ex] at (2.95,0.55) {\emph{exit:} grasp force\\settled};
\node[ex,fill=red!6,draw=red!45!black,rounded corners=1pt] at (4.95,0.55)
  {\emph{exit:} $\dot\theta>\epsilon_\theta$\\stall $\Rightarrow$ alert};
\node[ex] at (6.95,0.55) {\emph{exit:} $\theta_d$\\reached};
\node[ex] at (8.95,0.55) {\emph{exit:} gripper\\open};
\end{tikzpicture}}
\caption{Working-condition state machine. Each box gives state, phase
and (italic) control law; below it, the exit predicate ($\dot\theta$
door rate, $\epsilon_\theta$ breakaway threshold, $\theta_d$ target
angle). Red: C2 produces the latch transient the projection cannot
anticipate; green: C3 the pull that would move the furniture.}
\label{fig:fsm}
\end{figure}

Let $\hat{\bv{t}}=\hat{\bv{a}}\times \bv{r}_{o\to \bv{H}}/\|\bv{r}_{o\to \bv{H}}\|$ be the handle
tangent, the direction doing work on the door, and split the
measured wrench into $\bv{F}_t=(\bv{F}_h^{\T}\hat{\bv{t}})\hat{\bv{t}}$ and
$\bv{F}_r=\bv{F}_h-\bv{F}_t$. The controller is the five-state machine of
Fig.~\ref{fig:fsm}.

In detail: C0 is an admittance approach; C1 regulates the grasp
force against the load; C2 ramps the tangential force until the leaf
moves, its peak giving the stiction estimate
$\hat\tau_{\mathrm{bw}}$ used by Remark~\ref{rem:feas}, a stall
triggering the haptic alert; C4 damps the leaf and retracts along the
handle normal. In C3, with $\tilde\theta=\theta-\theta_d$,
$s=\dot{\tilde\theta}+\lambda\tilde\theta$ and regressor
$\bv{Y}=(\ddot\theta_d,\dot\theta_d,\sgn\dot\theta_d,\cos\theta_d,1)$,
\begin{equation}
\tau_h \;=\;\bv{Y}\,\hat{\bv{\beta}}\;+\;K_D\,s ,\qquad
\dot{\hat{\bv{\beta}}}\;=\;-\bv{\Gamma}\,\bv{Y}^{\T}s ,
\label{eq:swing}
\end{equation}
$\bv{\beta}=(I_h,b_v,\tau_c,\tau_g,\tau_0)^\T$ with
$\tau_g=-m_d g r_d\sin\gamma$ the gravity torque due to the hinge
tilt $\gamma$, read off $\hat{\bv{a}}$; $(b_v,\tau_c)$ are seeded from
$\hat\tau_{\mathrm{bw}}$, $\hat{\bv{\beta}}$ is kept in a box by projection,
and the grasp servo of C1 stays active. Unlike a fixed-gain PD,
\eqref{eq:swing} comes with a bound.

\begin{lemma}[Tracking bound of the adaptive swing]
\label{lem:track}
Let the leaf obey $I_h\ddot\theta+b_v\dot\theta
+\tau_c\sgn\dot\theta+\tau_g\cos\theta+\tau_0=\tau_h+\tilde\tau$
with $|\tilde\tau|\le\delta_\tau$ covering seal and latch friction
absent from $\bv{Y}$. Then ~\eqref{eq:swing} gives
$\limsup_t|s|\le\delta_\tau/K_D$ and
$|\tilde\theta|\le\delta_\tau/(\lambda K_D)$: the error is set by the
\emph{feedforward residual}, not by the load. Removing the
feedforward instead requires
$K_p\ge\tau_{\mathrm{req}}/\tilde\theta_{\mathrm{tol}}$.
\end{lemma}

\begin{IEEEproof}
With $I_h\dot s=-K_Ds+\bv{Y}\tilde{\bv{\beta}}+\tilde\tau$ and
$V=\tfrac12 I_h s^2+\tfrac12\tilde{\bv{\beta}}^\T\bv{\Gamma}^{-1}\tilde{\bv{\beta}}$,
the update cancels the parameter term and $\dot V\le0$ whenever
$|s|>\delta_\tau/K_D$.
\end{IEEEproof}

\section{BILATERAL FORCE FEEDBACK AND TRANSPARENCY}
\label{sec:transparency}

\subsection{Master--Slave Dynamics and the Force-Reflection Channel}
Both ports are Euler--Lagrange in their own coordinates, $\bv{p}_m$ for
the master and $\bv{x}_s$ for the slave in task space:
\begin{align}
\mat{M}_m\ddot{\bv{p}}_m+\mat{D}_m\dot{\bv{p}}_m &= \bv{f}_h+\bv{f}_m ,\label{eq:master}\\
\mat{M}_s\ddot{\bv{x}}_s+\mat{D}_s\dot{\bv{x}}_s &= J^{\T}\bigl(\bv{W}^{*}-\bv{f}_e\bigr) ,
\label{eq:slave}
\end{align}
with $\bv{f}_h$ the operator force, $\bv{f}_e$ the measured contact wrench and
$\mat{J}$ the handle Jacobian. The channel is \emph{position forward},
$\bv{x}_s^{d}=\kappa_p \mat{R}_{ms}^{\T}\bv{p}_m$, specialised to the amplitude
map \eqref{eq:amplitude}, with \emph{force back} through $\bv{f}_m$ of
\eqref{eq:fmaster}. Reduced to the mapped pull direction, all
impedances are SISO transfer functions along $\hat{\bv{e}}_m$, the
classical limit in the form we need.

\begin{lemma}[Displayed impedance of the force-reflection channel]
\label{lem:disp}
Let the slave track $\bv{x}_s^{d}$ with closed-loop impedance $Z_s$
(force per unit position error) against an environment $Z_e$, and let
$Z_m=\mat{M}_m s+\mat{D}_m$ be the master impedance.
Then the impedance the operator feels is
\begin{equation}
Z_{\mathrm{disp}}\;=\;Z_m\;+\;\kappa_p\kappa_f\,
\frac{Z_s\,Z_e}{Z_s+Z_e} ,
\label{eq:zdisp}
\end{equation}
so the channel is transparent, $Z_{\mathrm{disp}}\to\kappa_p\kappa_f
Z_e$, iff $\kappa_p\kappa_f=1$, $Z_m\!\ll\!Z_e$ and
$Z_s\!\gg\!Z_e$, and the relative defect is $Z_e/(Z_s+Z_e)$.
\end{lemma}

\begin{IEEEproof}
Eliminate $\dot{\bv{x}}_s$ from
$\bv{f}_e=Z_s(\dot{\bv{x}}_s^{d}-\dot{\bv{x}}_s)=Z_e\dot{\bv{x}}_s$ and
insert $\dot{\bv{x}}_s^{d}=\kappa_p\mat{R}_{ms}^{\T}\dot{\bv{p}}_m$,
$\bv{f}_m=\kappa_f\mat{R}_{ms}\bv{f}_e$ into~\eqref{eq:master}.
\end{IEEEproof}

The master force is
\begin{equation}
\bv{f}_m \;=\; \kappa_f\,\mat{R}_{ms}\,\bv{f}_e^{\mathrm{lin}} \;+\;
\kappa_d\,\dot{\bv{p}}_s^{\mathrm{proj}} \;+\; \bv{f}_{\mathrm{vf}}
\;+\; \bv{f}_{\mathrm{pc}},
\label{eq:fmaster}
\end{equation}
with $\mat{R}_{ms}=\mat{R}_m^{\T}\mat{R}_s$, $\kappa_f$ the force scale,
$\bv{f}_{\mathrm{vf}}$ the guard~\eqref{eq:guard} and $\bv{f}_{\mathrm{pc}}$ the
passivity damping. Ideal transparency (up to $\kappa_f$) means
$\bv{f}_m=\kappa_f \mat{R}_{ms}\bv{f}_e^{\mathrm{ideal}}$, where $\bv{f}_e^{\mathrm{ideal}}$
is the wrench the environment would produce under the
\emph{unprojected} command.

Two consequences follow. (i) The scaling loss
($\kappa_p\kappa_f=0.56$, Table~\ref{tab:params}) is a
\emph{constant} error calibrated away, whereas the projection adds a
\emph{state-dependent} one (Prop.~\ref{prop:transparency}); only the
latter is reported as $\bv{e}_f$. (ii) \eqref{eq:zdisp} depends on
$\bv{W}^{*}$ only through $\bv{f}_e$, so the filter changes the input,
not the channel (Prop.~\ref{prop:passivity}).

\subsection{Transmission Error and Its Bound}

\begin{definition}[Transmission error]
\label{def:err}
$\bv{e}_f = \bv{f}_m-\kappa_f \mat{R}_{ms}\bv{f}_e^{\mathrm{ideal}}$; the empirical
transparency score is the fraction of task time with
$\|\bv{e}_f\|\le\epsilon_{\mathrm{tr}}$ for a perceptual threshold
$\epsilon_{\mathrm{tr}}$.
\end{definition}

\begin{proposition}[Transparency cost of constraint enforcement]
\label{prop:transparency}
Let the door--furniture environment be passive with bounded
incremental gain $L_e$ from commanded wrench deviation to contact
force deviation, i.e.\
$\|\bv{f}_e-\bv{f}_e^{\mathrm{ideal}}\|\le L_e\|\bv{W}_d-\bv{W}^{*}\|$. Then, with
$\bv{f}_{\mathrm{pc}}=0$,
\begin{equation}
\|\bv{e}_f\| \;\le\; \kappa_f L_e\,\|\bv{W}_d-\bv{W}^{*}\|_{\bv{\Lambda}}
\;+\; k_g\,\bigl[g(\bv{W}_d)\bigr]_{+}^{2}
\;+\; \kappa_d\|\dot{\bv{p}}_s^{\mathrm{proj}}\| ,
\label{eq:bound}
\end{equation}
and $\|\bv{W}_d-\bv{W}^{*}\|=0$ whenever $\bv{W}_d\in\hat\cF(\eta)$: transparency
is exact up to damping, the guard being nonzero only in the band
$\dist(\bv{W}_d,\partial\hat\cF)\in(0,\ F_{\mathrm{ref}}\sqrt{\epsilon/k_g}]$,
where it grows quadratically.
\end{proposition}

\begin{IEEEproof}
Insert~\eqref{eq:fmaster} into Definition~\ref{def:err}, bound
$\bv{f}_e-\bv{f}_e^{\mathrm{ideal}}$ by $L_e$ and apply the triangle
inequality; inside $\hat\cF(\eta)$, $\bv{W}^{*}=\bv{W}_d$ and
$[g]_+=0$.
\end{IEEEproof}

\begin{lemma}[Radial form of the projection error]
\label{lem:radial}
If $\bv{W}_d$ violates exactly one constraint defining $\hat\cF(\eta)$, that
is the disc $\|\bv{F}_t\|\le c_t$ or a half-space $\langle\hat{\bv{n}},\bv{W}\rangle\le b$
(an edge margin, or the jamming ceiling), then
\begin{equation}
\bv{W}^{*}-\bv{W}_d\;=\;
\begin{cases}
-\bigl[\|\bv{F}_t^{d}\|-c_t\bigr]_{+}\dfrac{(\bv{F}_t^{d},0)}{\|\bv{F}_t^{d}\|},
& \text{disc},\\[2mm]
-\dfrac{\bigl[\langle\hat{\bv{n}},\bv{W}_d\rangle-b\bigr]_{+}}
{\|\hat{\bv{n}}\|_{\bv{\Lambda}^{-1}}^{2}}\,\bv{\Lambda}^{-1}\hat{\bv{n}} ,
& \text{half-space},
\end{cases}
\label{eq:radial}
\end{equation}
i.e.\ the error \emph{is} the excess, along the $\bv{\Lambda}$-normal of
the violated set; with $j$ active constraints cyclic projection gives
at most the sum of the $j$ excesses, which is what the guard \eqref{eq:guard}
renders.
\end{lemma}

Thus $\eta$ parametrizes the displacement--error frontier measured
in Sec.~\ref{sec:exp}.

\subsection{Passivity of the Constrained Bilateral Loop}
\label{subsec:pass}
\emph{Passivity.} Contact transitions inject energy, so the master
side is wrapped in a time-domain passivity
controller~\cite{ryu2004}: $P(k)=P(k-1)+\bv{f}_m^{\T}(k)\Delta \bv{p}_m(k)$ and
$\bv{f}_{\mathrm{pc}}=\alpha\dot{\bv{p}}_m$,
$\alpha\ge-P(k)/(\dot{\bv{p}}_m^{\T}\dot p_mT_s)$ at $T_s=1\,\mathrm{ms}$
whenever $P(k)<-\epsilon$, while the guard~\eqref{eq:guard} is the
gradient of a $C^{1}$ potential and hence conservative.

\begin{proposition}[Closed-loop passivity with the constraint]
\label{prop:passivity}
Let the environment be passive with incremental gain $L_e$ as in
Prop.~\ref{prop:transparency} and let the master port be wrapped in
the TDPA above. Then (a) the projection read as a static operator
$\bv{W}_d\mapsto \bv{W}^{*}$ is passive; (b) neither the energy the
master port can receive nor the damping the TDPA must inject exceeds
its unconstrained value; (c) the closed loop is passive.
\end{proposition}

\begin{IEEEproof}
(a) is Lemma~\ref{lem:proj}(iii) integrated in time; (b) follows from
Lemma~\ref{lem:proj}(i) and $\|\bv{f}_e\|\le L_e\|\bv{W}_d\|_{\bv{\Lambda}}$;
(c) is the passivity theorem for interconnected passive subsystems.
\end{IEEEproof}

The constraint thus costs fidelity, never passivity. The channel is
local ($250\,\mathrm{Hz}$), so the experiments do not exercise
Prop.~\ref{prop:passivity}; it certifies composition with TDPA
\emph{under} delay.

\subsection{The Converse: How Much Authority Must Be Paid For}

\begin{theorem}[Converse bound for memoryless safe filters]
\label{thm:converse}
Let $\Pi$ be any \emph{memoryless} command map
$\Pi(\bv{W}_d;\hat\mu,\hat m,\delta_\mu,\delta_m,\Delta)$ guaranteeing
Prop.~\ref{prop:static} for every $(\mu,m)$ in the box of
Assumption~\ref{as:param}, and let $\bv{W}_d$ produce no spin or tipping
demand with $N=\underline m g-F_z^{d}>0$. Then for every
$\eta\in(0,1]$, $\delta_\mu>0$ there are admissible instances in
which $\bv{W}_d\in\cF$, safe under the true parameters, yet
\begin{equation}
\bigl\|\bv{F}_t^{d}-\bv{F}_t^{*}\bigr\|\;\ge\;
(1-\eta)\,\mu\,N\;+\;2\eta\,\delta_\mu\,N\;+\;\Delta .
\label{eq:converse}
\end{equation}
\end{theorem}

\begin{IEEEproof}
Take $\mu=\hat\mu+\delta_\mu$ at the top of the box and
$\bv{F}_t^{d}=\mu N\hat{\bv{t}}$, other components zero; by the sufficiency part
of Lemma~\ref{lem:cwc} (CoP at $\bv{c}_0$, no residual spin) $\bv{W}_d\in\cF$.
The instance $\mu'=\mu-2\delta_\mu$ lies in the same box, yet at rest
the two are indistinguishable, because $\mu$ enters the observables only
through the sliding boundary, which is not reached. Being memoryless,
$\Pi$ must return a point of $\hat\cF(\eta)$, i.e.\
$\|\bv{F}_t^{*}\|\le\eta(\mu-2\delta_\mu)N-\Delta$, and
$\|\bv{F}_t^{d}-\bv{F}_t^{*}\|\ge\|\bv{F}_t^{d}\|-\|\bv{F}_t^{*}\|$ gives
\eqref{eq:converse}.
\end{IEEEproof}

\begin{corollary}[RFCP attains the converse bound]
\label{cor:tight}
Theorem~\ref{thm:converse} cannot be improved: with
$\mu=\hat\mu+\delta_\mu$ and $\bv{F}_t^{d}=\mu N\hat{\bv{t}}$, the radial
projection of Lemma~\ref{lem:radial} returns
$\|\bv{F}_t^{*}\|=\eta(\mu-2\delta_\mu)N-\Delta$, so \eqref{eq:converse}
holds \emph{with equality}, so RFCP is optimal among memoryless maps.
\end{corollary}

The loss splits into a \emph{stability-factor} term $(1-\eta)\mu N$
and an \emph{ignorance} term $2\eta\delta_\mu N$; no better optimiser
removes either. At $\eta=0.7$, $N\approx118\,\mathrm{N}$ (C1),
shrinking $\delta_\mu$ by $0.05$ returns $\approx\!8\,\mathrm{N}$ of
pull. The only escape is a non-memoryless $\Pi$ conditioned on
pressure-distribution measurements, for which ~\eqref{eq:converse} is
the floor.

\section{EXPERIMENTS}
\label{sec:exp}
\subsection{Setup, Identification and Protocol}
The slave is a UR10 with a parallel gripper and a wrist 6-axis F/T
sensor ($0.1\,\mathrm{N}$), gravity- and zero-drift calibrated over
several arm postures~\cite{zhou2022}; the master is a Phantom
($\approx8\,\mathrm{N}$ continuous). Both loops run at $1\,\mathrm{kHz}$,
the channel at $250\,\mathrm{Hz}$; an RGB-D
camera tracks door and body to $\approx0.5\,\mathrm{mm}$. The cabinets
stand unsecured \emph{on a table}, feet on wood or laminate. A
closed-leaf force ramp exposed a second failure mode: above
$F_{\mathrm{jam}}\approx30\,\mathrm{N}$ the leaf \emph{binds} in its
frame and the cabinet shifts, $F_{\mathrm{jam}}$ being the handle force
at which the body marker first moves while the door angle stays zero.

\emph{Identification} ($\approx\!8\,\mathrm{s}$ per instance): the
hinge $(\bv{o},\hat{\bv{a}})$ is fitted from a few degrees of damped rotation,
$\hat m,\hat\mu$ from a slow horizontal force ramp~\cite{zhou2016,teng2025}
(the force at first motion gives $\hat\mu\hat m g$, the pre-motion
load $\hat m$); three surfaces give $\delta_\mu\le0.05$,
$\delta_m\le0.6\,\mathrm{kg}$. Footprint and handle offset are
measured once per instance. On C1 the same ramps give
$\delta_W+\delta_{\mathrm{kin}}=0.6\,\mathrm{N}$ and
$\varepsilon_{\mathrm{dyn}}=1.4\,\mathrm{N}$, hence $\Delta=2\,\mathrm{N}$,
and $\bar F_{\mathrm{jam}}=26\,\mathrm{N}$, $10\%$ below the measured
$F_{\mathrm{jam}}$.
\label{sec:id}

\begin{table}[t]
\centering
\caption{Controller parameters (nominal).}
\label{tab:params}
\scriptsize
\begin{tabular}{@{}lclc@{}}
\toprule
Symbol & Value & Symbol & Value \\
\midrule
$\kappa_p$ & $1.6$ & $\kappa_f$ & $0.35$ \\
$r=\rho/2$ & $0.18\,\mathrm{m}$ (C1) & $\Delta$ & $2.0\,\mathrm{N}$ \\
$k_\theta$ & $12.5\,\mathrm{deg/cm}$ & $\theta_{\max}$ & $95^\circ$ \\
$K_D$ & $0.15\,\mathrm{Nm\,s/rad}$ & $\lambda$ & $4\,\mathrm{s^{-1}}$ \\
$k_g$ & $3.0\,\mathrm{N}$ & $\eta$ & $0.7$ (nominal) \\
$\delta_a,\delta_o$ & $1^\circ,\,2\,\mathrm{mm}$ & $\epsilon_{\mathrm{tr}}$ & $0.5\,\mathrm{N}$ \\
$\dot\theta_{\max},\ddot\theta_{\max}$ & $25^\circ/\mathrm{s},\,30^\circ/\mathrm{s}^2$ & $k_{\mathrm{st}}$ & $400\,\mathrm{N/m}$ \\
\bottomrule
\end{tabular}
\end{table}

\emph{Protocol.} Independent variables are \emph{furniture} (C1:
$12\,\mathrm{kg}$, $d_e=0.28\,\mathrm{m}$, $h=0.80\,\mathrm{m}$,
tipping-dominated; C2: $20\,\mathrm{kg}$, $h=0.60\,\mathrm{m}$,
sliding-dominated, handle $0.28\,\mathrm{m}$ off-axis; C3:
$7\,\mathrm{kg}$, on the feasibility boundary) and \emph{support}
(wood $\mu\approx0.47$, laminate
$\mu\approx0.28$). Conditions: P ($\eta=0.7$); NG ($k_g=0$); NP
(no projection); POS. Five operators (4 novice) ran 5 trials per
condition in Latin-square order on C1 over wood; further trials on
C2, C3 and the laminate support test generalization.
Metrics: displacement, $\|\bv{e}_f\|$, peak force, times, angle
RMSE, success, NASA-TLX~\cite{hertzum2021}; counts use
Bonferroni-corrected Fisher exact tests. All $100$ trials ($25$ per
condition) are physical measurements.

\subsection{Results}
\label{sec:results}

\begin{figure}[t]
\centering
\includegraphics[width=0.48\columnwidth]{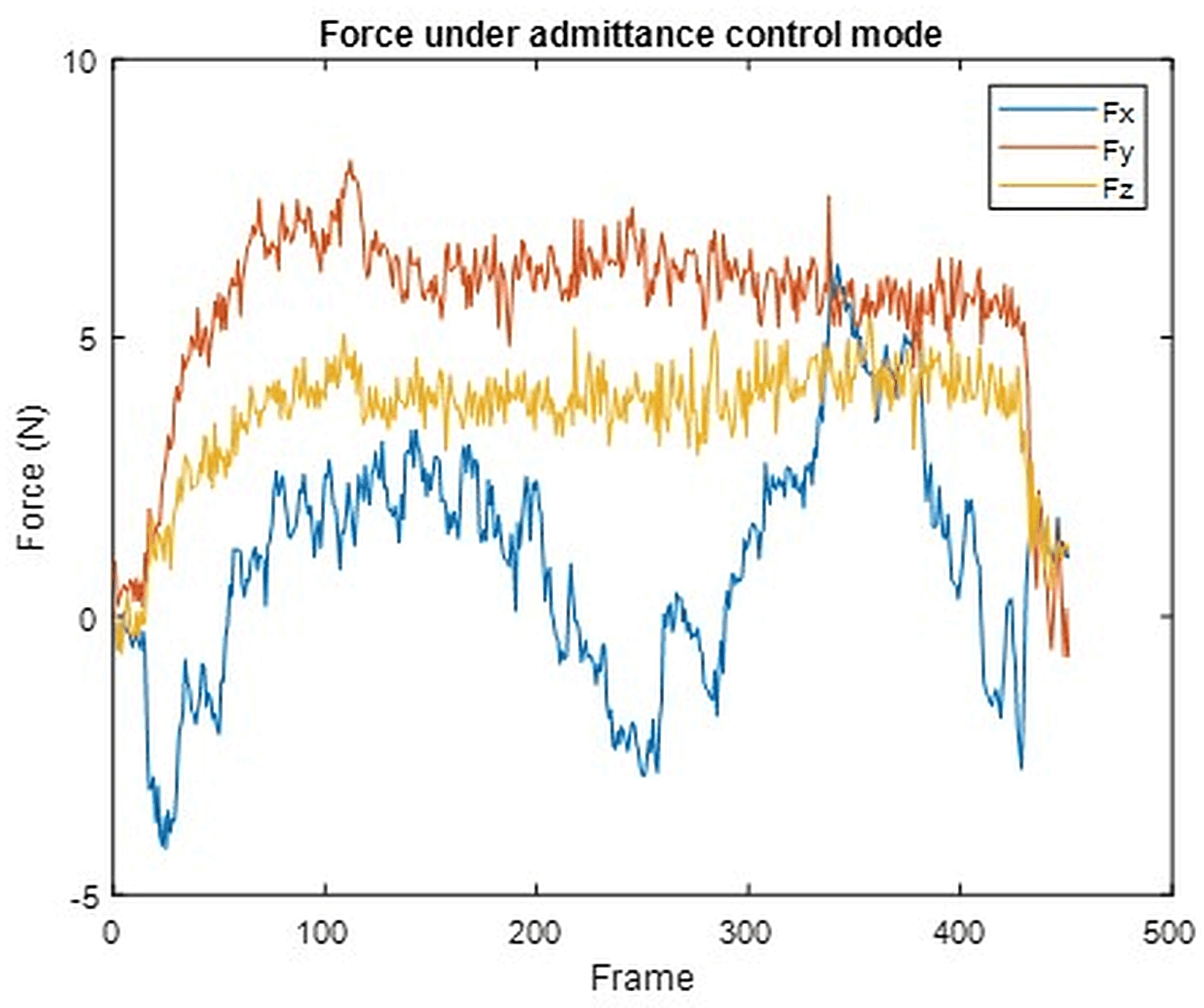}%
\hfil
\includegraphics[width=0.48\columnwidth]{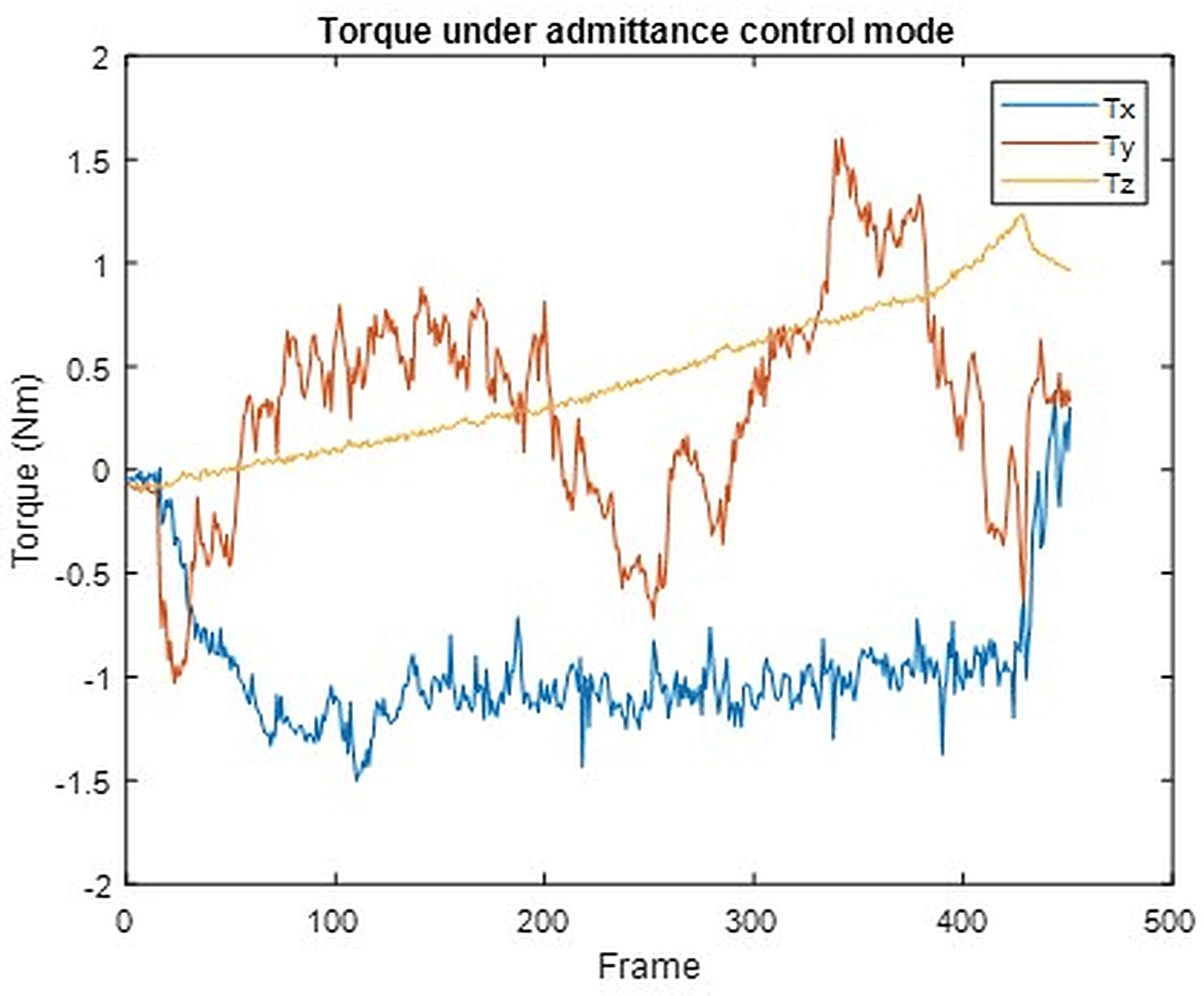}
\caption{Measured wrist F/T in one admittance-mode trial: force
$F_{x,y,z}$ (left) and moment $T_{x,y,z}$ (right) in the sensor frame
versus sample frame.}
\label{fig:wrench}
\end{figure}

\begin{table}[t]
\centering
\caption{Task metrics, mean $\pm$ s.d.\ over 5 operators
$\times$ 5 trials, furniture C1 on wood.}
\label{tab:results}
\scriptsize
\setlength{\tabcolsep}{4.8pt}
\begin{tabular}{@{}lcccc@{}}
\toprule
Metric & P & NG & NP & POS \\
\midrule
Disp.\ (mm) & $0.9\pm0.4$ & $6.8\pm2.3$ & $4.1\pm1.9$ & $9.4\pm3.1$ \\
Viol.\ $>5$ mm & 0/25 & 5/25 & 3/25 & 8/25 \\
Peak $F$ (N) & $20.8\pm2.0$ & $27.4\pm3.1$ & $25.1\pm2.8$ & $30.6\pm4.9$ \\
Error (N) & $0.41\pm0.13$ & $0.28\pm0.09$ & $0.33\pm0.11$ & $0.62\pm0.21$ \\
$t_{\mathrm{bw}}$ (s) & $3.4\pm0.7$ & $3.1\pm0.6$ & $3.3\pm0.7$ & $4.2\pm1.3$ \\
$t_{\mathrm{task}}$ (s) & $14.6\pm3.1$ & $16.2\pm3.6$ & $15.4\pm3.4$ & $19.1\pm6.0$ \\
Success (\%) & 96.0 & 84.0 & 88.0 & 72.0 \\
RMSE/TLX & $2.7^\circ$, $40\pm9$ & $3.1^\circ$, $50\pm11$ & $2.9^\circ$, $47\pm10$ & $4.8^\circ$, $61\pm13$ \\
\bottomrule
\end{tabular}
\end{table}

\begin{figure}[t]
\centering
\includegraphics[width=0.66\columnwidth]{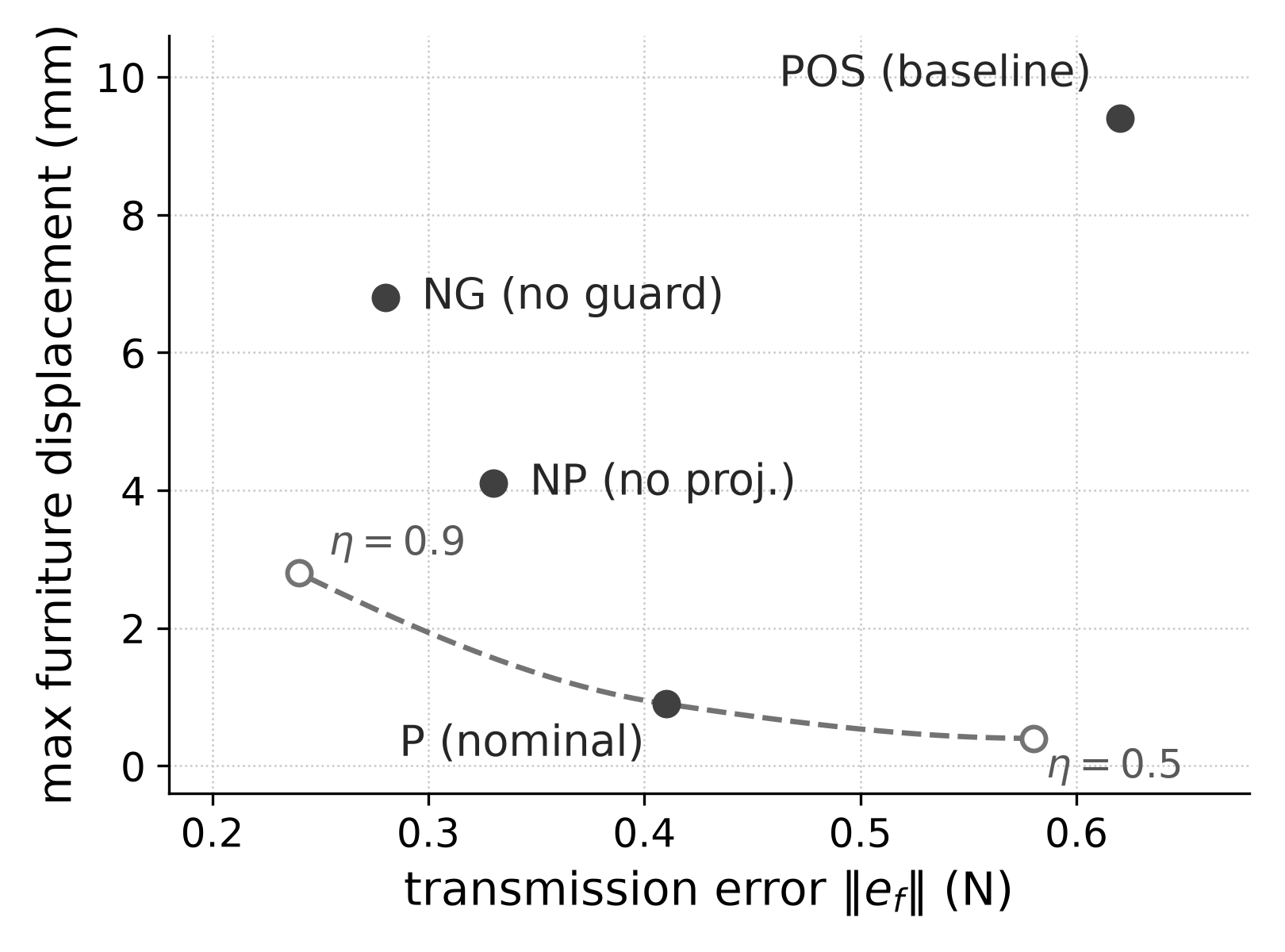}
\caption{Stability--transparency trade-off: maximum furniture
displacement versus transmission error $\|\bv{e}_f\|$. Filled markers:
measured conditions P, NG, NP, POS ($25$ trials each); dashed: analytic
frontier of Prop.~\ref{prop:transparency} at the identified
$\hat\mu,\hat m$, open markers at $\eta=0.9$ (left) and $0.5$ (right).}
\label{fig:pareto}
\end{figure}

\emph{Furniture stability.} With projection and guard active,
displacement was $0.9\pm0.4\,\mathrm{mm}$ and no trial exceeded the
$5\,\mathrm{mm}$ abort threshold,
with peak force at $20.8\pm2.0\,\mathrm{N}$, below the
$\bar F_{\mathrm{jam}}=26\,\mathrm{N}$ clamp. POS is the only
condition whose mean peak exceeds the measured
$F_{\mathrm{jam}}\approx30\,\mathrm{N}$; it has most violations
($8/25$) and lowest success ($72\%$; P vs.\ POS, Fisher
$p=0.004$). Removing the guard (NG) gave $5/25$ violations on C1 and
degraded further on the laminate support; removing the projection
(NP) $3/25$. The two mechanisms are complements, not substitutes: the
projection enforces ~\eqref{eq:robust} \emph{within} the envelope of
Assumptions~\ref{as:qs}--\ref{as:param} and cannot anticipate the
latch-breakaway transient. Its signature is the NG peak force
($27.4\pm3.1\,\mathrm{N}$) exceeding the clamp, while the guard keeps
the operator inside that envelope. Only the full P attains $0/25$;
after Bonferroni correction only P vs.\ POS is significant
($p=0.012$). In Fig.~\ref{fig:pareto} the ablations sit off the
analytic frontier, which passes through the measured P point.

\emph{Arbitration in shared-control terms.} The binding cap on C1 is
$\bar c=\min\{\eta\underline\mu\underline m g,\ \eta\underline m
g\,d_e/h,\ \bar F_{\mathrm{jam}}\}-\Delta=26\,\mathrm{N}$, and the peak
demand reached $0.80$ (P), $0.97$ (NP), $1.05$ (NG) and $1.18$ (POS)
of $\bar c$: under P the arbitrator (Lemma~\ref{lem:arb}) stayed
idle.

\emph{Measured wrench} (Fig.~\ref{fig:wrench}). Force peaks at
$\approx\!8\,\mathrm{N}$ at the latch; a $1.5\,\mathrm{Nm}$ $T_y$
spike marks breakaway.

\emph{Which constraint binds.} On C1 tipping bound throughout (spin
under $5\%$), as ~\eqref{eq:tipping} predicts for a tall body; on C2
the off-axis handle made spin the clipped constraint, which a filter
bounding only $\|\bv{F}_t\|$ would miss. The failures (gripper slip,
two C3 stalls) were predicted by Remark~\ref{rem:feas}.

\subsection{Discussion}
\label{sec:disc}
Three claims are supported: RFCP keeps furniture static across masses,
geometries and supports; the transparency cost is bounded and below
the perceptual threshold at $\eta=0.7$; and the trade-off is
one-dimensional (Fig.~\ref{fig:pareto}).
Caveats: the NG/NP differences are counts, not confirmed
inferences, and $\Delta$ was calibrated on C1. Four open problems delimit the
claim: (O1) online friction identification, since $\delta_\mu\approx0.15$
on a dusty support consumes most of the margin; (O2) delay and packet loss, for which
Thm.~\ref{thm:converse} has no analogue; (O3) tuned virtual fixtures
and learned arbitrators, equated to RFCP analytically
(Lemma~\ref{lem:arb}) but not empirically; (O4) fast pulls and
multi-body scenes, which break Assumption~\ref{as:qs}.

\section{CONCLUSION}
We formalized and measured the trade-off between keeping unsecured
furniture still and telling the operator the truth. RFCP gives
staticity, a feasibility certificate, passivity, an a-priori
transmission-error bound and a converse that is tight among memoryless
filters. Over $100$ trials at $\eta=0.7$, displacement falls from
$9.4\pm3.1$ to $0.9\pm0.4\,\mathrm{mm}$, success rises from $72\%$ to
$96\%$ and workload drops, while transmission error stays under the
$0.5\,\mathrm{N}$ threshold. Because every input of the certificate is
identified at run time, one filter covered tipping-, spin- and
feasibility-limited cabinets on two supports, and deployment reduces
to choosing $\eta$. The converse shows that the remaining gap must be
paid for with information, such as online friction identification
(O1)--(O4).

\balance
\def\IEEEbibitemsep{0pt}

\end{document}